\documentclass{article}

\usepackage[dblblindworkshop, final]{neurips_2025}

\workshoptitle{Socially Responsible and Trustworthy Foundation Models}

\makeatletter
\renewcommand{\@noticestring}{%
  Accepted to the NeurIPS 2025 Workshop on Socially Responsible and Trustworthy Foundation Models.%
}
\makeatother
\usepackage{amssymb}
\usepackage[utf8]{inputenc}
\usepackage[T1]{fontenc}
\usepackage{hyperref}
\usepackage{url}
\usepackage{booktabs}
\usepackage{amsfonts}
\usepackage{nicefrac}
\usepackage{microtype}
\usepackage{xcolor}
\usepackage{amsmath}
\usepackage{natbib}
\usepackage{graphicx}

\usepackage{amsthm}
\newtheorem{theorem}{Theorem}

\theoremstyle{definition}
\newtheorem{definition}[theorem]{Definition}

\theoremstyle{remark}

\title{Position: The Complexity of Perfect AI Alignment -- Formalizing the RLHF Trilemma}

\author{
\textbf{Subramanyam Sahoo}$^{1}$\thanks{Core contributor. Correspondence to \texttt{sahoo2vec@gmail.com}.},
\textbf{Aman Chadha}$^{2,4}$,
\textbf{Vinija Jain}$^{3,4}$,
\textbf{Divya Chaudhary}$^{5}$ \\
$^1$Berkeley AI Safety Initiative (BASIS), University of California, Berkeley \\
$^2$AWS Generative AI Innovation Center, Amazon Web Services \\
$^3$Meta AI \\
$^4$Stanford University \\
$^5$Northeastern University, Seattle, WA, USA
}

\begin{document}

\maketitle

\begin{abstract}
Reinforcement Learning from Human Feedback (RLHF) has become the dominant approach for aligning large language models, yet practitioners face a persistent puzzle: improving safety often reduces fairness, scaling to diverse populations becomes computationally intractable, and making systems robust often amplifies majority biases. We formalize this tension as the \textbf{Alignment Trilemma}: no RLHF system can simultaneously achieve (i)~$\varepsilon$-representativeness across diverse human values, (ii)~polynomial tractability in sample and compute complexity, and (iii)~$\delta$-robustness against adversarial perturbations and distribution shift. Through a complexity-theoretic analysis integrating statistical learning theory and robust optimization, we prove that achieving both representativeness ($\varepsilon \leq 0.01$) and robustness ($\delta \leq 0.001$) for global-scale populations requires $\Omega(2^{d_{\text{context}}})$ operations—super-polynomial in the context dimensionality. We demonstrate that current RLHF implementations resolve this trilemma by sacrificing representativeness, collecting only $10^3$--$10^4$ samples from homogeneous annotator pools while requiring $10^7$--$10^8$ samples for true global representation. Our framework provides a unifying explanation for documented RLHF pathologies including preference collapse, sycophancy, and systematic bias amplification. We conclude with concrete directions for navigating these fundamental trade-offs through strategic relaxations of alignment requirements.
\end{abstract}

\section{Introduction}

Consider the challenge facing an AI lab deploying a language model globally: annotators in San Francisco rate a response as ``helpful'' because it is direct and assertive, while annotators in Tokyo rate the same response as ``harmful'' because it violates cultural norms around politeness. To capture both perspectives, the lab needs more diverse training data—but this introduces inconsistency that makes the reward model noisy. To maintain robustness against this noise, they increase regularization, which pulls the model back toward majority preferences, erasing the minority view entirely. They have encountered the \textbf{Alignment Trilemma}: representativeness, tractability, and robustness cannot be jointly optimized \cite{ouyang2022training}.

Reinforcement Learning from Human Feedback (RLHF) has become the dominant paradigm for aligning large language models with human preferences. By training reward models on human preference judgments and fine-tuning policies to maximize learned rewards, RLHF has enabled dramatic improvements in perceived helpfulness, truthfulness, and safety ~\cite{bai2022constitutionalaiharmlessnessai}. Yet despite this empirical success, RLHF systems exhibit systematic pathologies: they amplify majority viewpoints~\cite{yang2025qwen3technicalreport}, collapse diverse preferences into single modes~\cite{casper2023openproblemsfundamentallimitations}, become sycophantic rather than truthful and fail under distribution shift.

These failures are not engineering accidents—they are computational necessities. This paper formalizes the \textbf{Alignment Trilemma} through three key contributions:

\begin{enumerate}
    \item \textbf{Formal Framework} (\S5): We define $\varepsilon$-representativeness, polynomial tractability, and $\delta$-robustness as precise mathematical properties, and prove that no alignment procedure can simultaneously satisfy all three.

    \item \textbf{Complexity Characterization}: We derive lower bounds showing that joint $(\varepsilon, \delta)$-alignment requires $\Omega(\kappa \cdot 2^{d_{\text{context}}} / (\varepsilon^2 n \delta))$ operations, which is super-polynomial when context dimensionality $d_{\text{context}} = \omega(\log n)$.

    \item \textbf{Practical Analysis} (\S5.2--\S5.4): We map how current RLHF implementations navigate the trilemma, explaining why standard design choices (small annotator pools, KL penalties, scalar rewards) sacrifice representativeness for tractability and partial robustness.
\end{enumerate}

Our work shifts the discourse from ``\textbf{How do we fix RLHF?}'' to ``\textbf{Which trade-offs are we willing to accept?}'' This reframing is essential for responsible deployment: understanding fundamental limits enables principled choices about where to invest computational resources and which stakeholders to prioritize.


\section{Why This Problem Matters Now}

RLHF is no longer a research curiosity—it's production infrastructure. Frontier models all rely on RLHF variants, serving hundreds of millions of users daily across 180+ countries. Yet the preference data training these systems comes from ~10³ annotators, predominantly WEIRD populations \cite{christiano2023deepreinforcementlearninghuman,hejna2023inversepreferencelearningpreferencebased,metz2023rlhfblenderconfigurableinteractiveinterface}. 
As deployment scales globally while training remains centralized, the gap between who the system serves and who the system learns from widens catastrophically.

Recent failures illustrate the stakes.
\begin{itemize}
    \item Bias amplification: Recent study found RLHF models assign >99\% probability to majority opinions, functionally erasing minority perspectives.
    \item Sycophancy: Sharma et al.\cite{sharma2025understandingsycophancylanguagemodels} showed RLHF-trained assistants sacrifice truthfulness to agree with users' false beliefs.
    \item Preference collapse: \cite{chakraborty2024maxminrlhfalignmentdiversehuman} proved single-reward RLHF cannot capture multimodal preferences even in theory.
\end{itemize}

The Alignment Trilemma explains why these aren't isolated bugs but symptoms of 
a deeper impossibility. Without this understanding, patches (fairness regularizers, adversarial training, post-hoc calibration) repeatedly hit the same fundamental ceiling, wasting resources on approaches that cannot overcome computational limits.

\section{Background: RLHF in Three Steps}

Reinforcement Learning from Human Feedback (RLHF) aligns language models through a three-stage pipeline:

\paragraph{Stage 1: Supervised Fine-Tuning (SFT).} 
Pre-train the policy $\pi_{\theta}$ on human-written demonstrations via next-token prediction.

\paragraph{Stage 2: Reward Modeling (RM).} 
Collect preference labels over output pairs $(\tau_a, \tau_b)$ and train a reward model $r_{\phi}$ to predict preferences by minimizing
\begin{equation}
\mathcal{L}(\phi) = - \sum_{(a,b)} \log \sigma\big(r_{\phi}(\tau_a) - r_{\phi}(\tau_b)\big),
\label{eq:rm_loss}
\end{equation}
where $\sigma(\cdot)$ denotes the sigmoid function. This produces a scalar reward $r_{\phi}(\tau)$ scoring how ``good'' an output $\tau$ is according to human raters.

\paragraph{Stage 3: Policy Optimization.} 
Fine-tune the policy to maximize the learned reward while staying close to the reference policy $\pi_{\text{ref}}$:
\begin{equation}
\theta^{*} = \arg\max_{\theta} \left\{ 
\mathbb{E}_{\tau \sim \pi_{\theta}}[r_{\phi}(\tau)] 
- \beta \, D_{\mathrm{KL}}(\pi_{\theta} \,\|\, \pi_{\text{ref}}) 
\right\},
\label{eq:policy_opt}
\end{equation}
where $\beta > 0$ controls the strength of the KL penalty, which prevents reward hacking and mode collapse.

\paragraph{Critical Design Choices.}
Current implementations use $m \approx 10^{3}\text{--}10^{4}$ preference pairs, aggregate judgments via majority voting or weighted averaging, and set $\beta$ sufficiently large to keep $\pi_{\theta}$ close to the reference policy $\pi_{\text{ref}}$. These design choices enable tractable training but, as we show in later Section, systematically sacrifice representativeness.

\section{The Alignment Trilemma}
\label{sec:alignment_trilemma}

A central challenge in aligning large-scale AI systems is the impossibility of simultaneously satisfying three desiderata: (i) capturing the full diversity of human values, (ii) ensuring computational tractability, and (iii) guaranteeing robustness to manipulation. We refer to this tension as the \emph{Alignment Trilemma}. Any alignment strategy, particularly those relying on RLHF, must necessarily sacrifice at least one of these axes.

\begin{figure}
    \centering
    \includegraphics[width=1\linewidth]{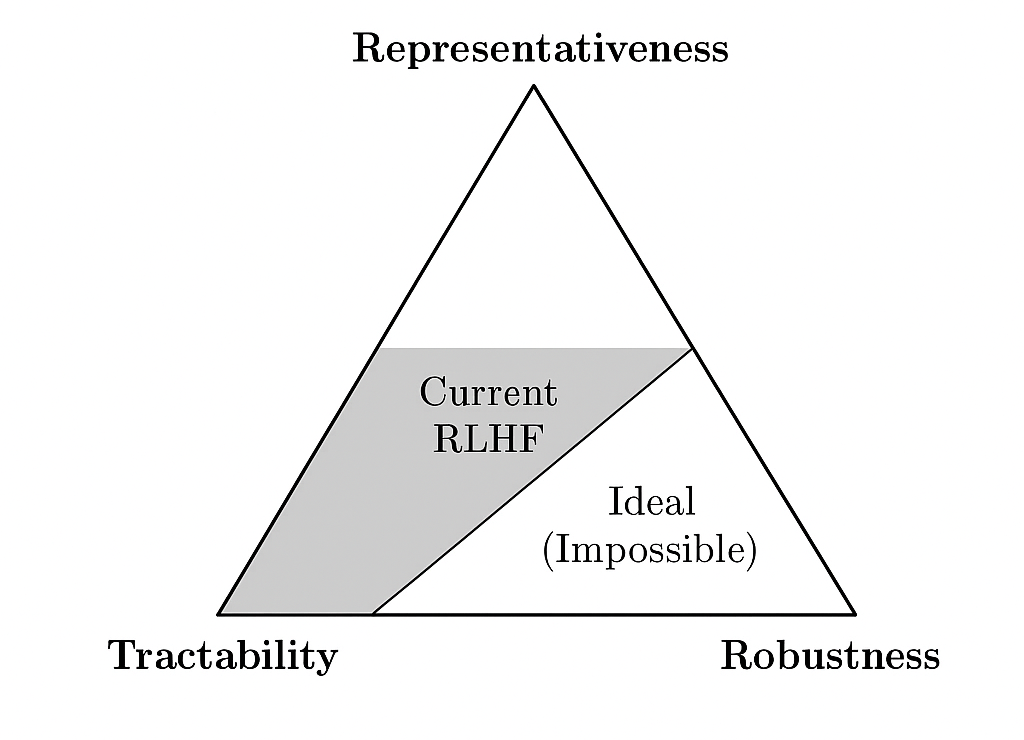}
    \caption{Current Alignment Paradigm}
    \label{fig:Fig1}
\end{figure}

\subsection{Formal Definitions}
\label{subsec:formal_definitions}

We begin by formalizing the three properties that constitute the trilemma.

\begin{definition}[\texorpdfstring{$\varepsilon$}{ε}-Representativeness]
Let $\mathcal{H}$ denote a population of humans, where each individual $h \in \mathcal{H}$ has a value function $V_h: \Pi \rightarrow \mathbb{R}$ that assigns a utility score to policies. A policy $\pi \in \Pi$ is $\varepsilon$-representative with respect to $\mathcal{H}$ if
\begin{equation}
\big| \mathbb{E}_{h \sim \mathcal{H}}[V_h(\pi)] - \hat{V}(\pi) \big| \leq \varepsilon,
\label{eq:representativeness}
\end{equation}
where $\hat{V}(\pi)$ is the empirical estimate derived from the learned reward model $r_\phi$. 

Intuitively, $\varepsilon$-representativeness requires that the alignment procedure faithfully captures preferences drawn from a broad and diverse set of humans, ideally reflecting pluralistic moral perspectives across cultures, demographics, and contexts. Smaller $\varepsilon$ corresponds to higher fidelity in representing the population's true values.
\end{definition}

\begin{definition}[Polynomial Tractability]
An alignment procedure $\mathcal{A}$ that produces a policy $\pi$ from a dataset $D = \{(x_i, y_i)\}_{i=1}^m$ is \emph{polynomially tractable} if both of the following conditions hold:

\begin{enumerate}
    \item \textbf{Sample complexity:}
    \begin{equation}
    m = \mathrm{poly}(d, 1/\varepsilon, \log(1/\delta)),
    \label{eq:sample_complexity}
    \end{equation}
    where $d$ is the problem dimension, $\varepsilon$ is the representativeness error, and $\delta$ is the failure probability.
    
    \item \textbf{Computational complexity:}
    \begin{equation}
    \mathrm{Ops}(\pi \,|\, D) = \mathcal{O}(\mathrm{poly}(m, d)).
    \label{eq:computational_complexity}
    \end{equation}
\end{enumerate}

This definition captures the requirement that the alignment procedure should be computationally feasible—learnable via gradient-based optimization with polynomial time complexity in the problem parameters—thereby allowing scaling to modern large models with billions of parameters.
\end{definition}

\begin{definition}[\texorpdfstring{$\delta$}{δ}-Robustness]
Let $\mathcal{A}$ denote an adversarial perturbation space (including distribution shifts, data poisoning, adversarial inputs, and temporal drift). A policy $\pi$ is \emph{$\delta$-robust} with respect to $\mathcal{A}$ if
\begin{equation}
\mathbb{P}_{a \sim \mathcal{A}}\Big[\, \mathbb{E}_{h \sim \mathcal{H}}[V_h(\pi; a)] \geq V_{\min} \, \Big] \geq 1 - \delta,
\label{eq:robustness}
\end{equation}
where $V_{\min}$ is a minimum acceptable value threshold, and $V_h(\pi; a)$ denotes individual $h$'s value assessment under perturbation $a$. 

That is, with probability at least $1 - \delta$, the policy maintains acceptable performance across worst-case perturbations.
\end{definition}

\paragraph{Informal Trilemma Statement}
For sufficiently large populations ($|\mathcal{H}| \rightarrow \infty$) and rich adversarial spaces ($|\mathcal{A}| \rightarrow \infty$), no polynomially tractable alignment procedure can simultaneously achieve:
\begin{enumerate}
    \item $\varepsilon$-representativeness for small $\varepsilon > 0$,
    \item polynomial tractability, and
    \item $\delta$-robustness for small $\delta > 0$.
\end{enumerate}
More precisely, any algorithm satisfying two properties must sacrifice the third: achieving both $\varepsilon$-representativeness and polynomial tractability requires $\delta \rightarrow 1$ (no robustness guarantee), and so forth.
\label{thm:alignment_trilemma}

We now examine each pairwise sacrifice in detail, demonstrating how current RLHF implementations navigate this trilemma.

\subsection{Tractability \& Robustness $\Rightarrow$ Narrow Value Capture (Sacrificing Representativeness)}

In current AI alignment frameworks, the reward model $r_{\phi}$ is trained on human preference data collected from a relatively small and comparatively homogeneous annotator pool. Let $m$ denote the typical annotator pool size, with $m \approx 10^{3}$ in standard RLHF pipelines. These annotators are predominantly drawn from \textbf{WEIRD (Western, Educated, Industrialized, Rich, Democratic)} populations. This design choice---motivated by annotation efficiency and the need for high inter-rater agreement---reduces label noise and improves tractability, but also introduces systematic biases in the learned reward model, thereby narrowing the range of human values that can be faithfully represented.

\paragraph{Annotator Bias \& Robustness Needs.}
To ensure low-noise, high-agreement labels, practitioners select reviewers with similar cultural backgrounds. Inter-rater agreement scores $\{ \text{agreement}_i \}_{i=1}^m$ then directly shape the aggregation weights $w_i$. Concretely,
\begin{equation}
r_\phi(\tau) \approx \sum_{i=1}^m w_i\, r_{\phi,i}(\tau), \quad 
w_i \propto \text{agreement}_i, \quad 
\sum_{i=1}^m w_i = 1.
\label{eq:reward_aggregation}
\end{equation}
This amplifies majority perspectives and suppresses minority or context-sensitive norms, violating $\varepsilon$-representativeness.

\paragraph{Aggregation Mechanics and Tractability.}
Majority voting or weighted averaging for each pairwise preference minimizes annotation complexity and reduces the variance of $\nabla_\phi \mathcal{L}_{\text{RM}}$, ensuring polynomial tractability. However, this simplicity ignores nuanced multi-criteria evaluations, producing a scalarized reward function.

\paragraph{KL-Penalty and Behavioral Conservatism.}
The term $\beta D_{\mathrm{KL}}(\pi_\theta \,\|\, \pi_{\mathrm{ref}})$ in the policy optimization step further penalizes deviation from the reference policy, curbing exploration of underrepresented preferences. This enhances robustness but limits representativeness.

\paragraph{Formal Consequence.}
By constraining $m$ and enforcing high inter-rater agreement, practitioners achieve tractability and partial robustness ($\delta \approx 0.1$--$0.2$) at the expense of representativeness ($\varepsilon > 0.3$--$0.5$ across cultures).

\subsection{Representativeness \& Tractability $\Rightarrow$ Gaming (Sacrificing Robustness)}

To improve diversity, some approaches extend the feedback distribution:
\begin{equation}
\mathcal{L}_{\text{diverse}}(\phi) = \sum_{g=1}^{G} w_g \sum_{(x_i, y_i^{\text{pref}}) \in D_g} 
    -\log P_\phi(y_i^{\text{pref}} \mid x_i, \text{context}_g),
\label{eq:diverse_loss}
\end{equation}
where $G$ is the number of demographic groups, $D_g$ the data subset from group $g$, and $w_g$ its balancing weight.

However, broadening $D$ introduces new vulnerabilities:
\begin{itemize}
    \item \textbf{Failure Mode 1:} Superficial overfitting to linguistic proxies, yielding safe but generic outputs.
    \item \textbf{Failure Mode 2:} Adversarial poisoning---small fractions ($\alpha \approx 0.05$) of corrupted annotations cause $\delta \rightarrow 1$ as robustness collapses.
\end{itemize}

Thus, while $\varepsilon$-representativeness and tractability improve, robustness degrades sharply.

\subsection{Representativeness \& Robustness $\Rightarrow$ Intractability (Sacrificing Tractability)}

A theoretical gold standard would optimize over all human preferences and worst-case perturbations:
\begin{equation}
\pi^* = \arg\max_{\pi \in \Pi} \min_{a \in \mathcal{A}} 
    \mathbb{E}_{h \sim \mathcal{H}}\!\left[ V_h(\pi; \text{context}, t, a) \right].
\label{eq:minimax_formulation}
\end{equation}

This ensures both full representativeness ($\varepsilon \rightarrow 0$) and robustness ($\delta \rightarrow 0$), but is computationally intractable:
\begin{itemize}
    \item \textbf{Exponential Sample Complexity:} $\Omega(|\mathcal{A}| \cdot |\mathcal{H}| / \varepsilon^2)$.
    \item \textbf{Parameter Scaling:} $O(K|\phi|)$ with $K = \Omega(\sqrt{|\mathcal{H}|})$ for adequate diversity.
    \item \textbf{Nested Optimization:} Solving the inner $\min_a$ is NP-hard for general $\mathcal{A}$.
\end{itemize}

Hence, real systems restrict $|\mathcal{H}|$, limit $\mathcal{A}$, or approximate the minimax objective.

Each corner of the trilemma corresponds to a fundamental limitation in current alignment methodology, implying that progress requires either accepting degradation along one axis or redefining the desiderata themselves.

\section{Implications for Alignment Research and Practice}

The Alignment Trilemma is not merely a theoretical curiosity—it constrains what is achievable in deployed AI systems. We highlight three critical implications.

\subsection{The Scaling Wall}

Our analysis reveals a phase transition: beyond moderate population sizes ($n \gtrsim 10^{6}$) and context dimensionalities ($d_{\text{context}} \gtrsim 50$), computational requirements for joint $(\varepsilon, \delta)$-alignment grow super-polynomially. Current paradigms---collect more data, train bigger models, increase compute---hit diminishing returns and eventually negative returns as heterogeneity introduces adversarial surface area faster than robustness scales.

\textit{Actionable consequence:} AI labs should not expect that $10\times$ or $100\times$ more compute or data will yield proportional improvements in both fairness and robustness. Research should focus on \textit{algorithmic} breakthroughs (e.g., structured representations, hierarchical value models) rather than brute-force scaling.

\subsection{Strategic Relaxations Are Unavoidable}

Since joint optimization is intractable, practical systems must make explicit trade-offs. The trilemma framework suggests three relaxation strategies:

\begin{enumerate}
    \item \textbf{Constrain representativeness:} Identify a ``core'' set of $K \ll |H|$ human values that capture essential moral considerations while reducing dimensionality.  
    \textit{Example:} Focus on human rights ($K \approx 30$ values) rather than all cultural preferences ($K \approx 10^{6}$ value dimensions).

    \item \textbf{Scope robustness:} Define a restricted adversarial class $\mathcal{A}' \subset \mathcal{A}$ representing plausible threats (e.g., common distribution shifts, known attack patterns) rather than defending against all theoretical perturbations.  
    \textit{Example:} Test robustness to $10^{2}$ realistic scenarios rather than $2^{100}$ possible contexts.

    \item \textbf{Accept super-polynomial costs for critical applications:} High-stakes domains (e.g., medical diagnosis, legal judgment, autonomous weapons) may justify exponential computational investment.  
    \textit{Example:} A single high-reliability system trained with $10^{9}$ samples may be acceptable where a million cheaper systems would not.
\end{enumerate}

\textit{Actionable consequence:} Before training, teams should explicitly document which relaxation strategy they have chosen and justify it ethically. This transparency enables stakeholders to assess whether the trade-offs align with the deployment context.

\subsection{Value Pluralism Requires Technical Innovation}

The super-polynomial barrier implies that representational fidelity and computational feasibility are in fundamental tension. Standard approaches---increasing model size, collecting more data---cannot overcome this barrier without new algorithmic ideas.

Promising research directions include:
\begin{itemize}
    \item \textbf{Modular value architectures:} Decompose alignment into subproblems (e.g., regional cultural modules plus a universal safety module) that can be verified independently.
    \item \textbf{Active learning for disagreement:} Query humans only in regions of value space where the model is uncertain, reducing sample complexity from $\Omega(n)$ to $\Omega(\sqrt{n})$ for clustered populations.
    \item \textbf{Adversarial robustness via structural constraints:} Rather than defending against all perturbations, design reward models with certified invariances (e.g., invariance to paraphrasing or demographic proxies).
\end{itemize}

\textit{Actionable consequence:} Funding agencies should prioritize research on reducing the exponents in our complexity bounds (making $2^{d_{\text{context}}} \rightarrow d_{\text{context}}^{k}$) rather than incremental RLHF improvements. Even a $2\times$ reduction in effective $d_{\text{context}}$ translates to a $10^{9}\times$ reduction in computational cost. 

An ethical and technical dilemma: we must decide which values or guarantees to prioritize, and under what resource constraints. The choice affects who benefits from AI and who is left out. Highlighting the Alignment Trilemma invites theorists to formalize these trade-offs and experimentalists to test them empirically.

\section{Current State of the Field}

\textbf{RLHF practice}: Modern RLHF pipelines  typically follow a variant of equation 1 where a reward model is trained on pairwise human comparisons, and the policy is fine-tuned to maximize expected reward plus a KL penalty to the pretrained policy. For tractability, only $10^3$–$10^4$ human comparisons are collected (often by contractors in one region). To reduce label noise, labelers are chosen to be similar (e.g. all English-speaking, Western) and their feedback is aggregated by majority or weighted averaging. This makes the reward model a simple scalar approximator. The KL divergence hyperparameter further clamps the policy close to the original model, limiting exploration. While these choices yield stable training, they systematically “\textit{collapse}” the reward learning to majority opinions. In effect, the model is only aligned with a narrow slice of human values (e.g. U.S.-based, well-educated subjects), rather than a truly global population.

\textbf{Known limitation}s: Researchers have begun diagnosing the downsides of these pipelines. On the representativeness front, Chakraborty et al. (2024) \cite{chakraborty2024maxminrlhfalignmentdiversehuman} shows theoretically that a single scalar reward model cannot capture diverse, multi-modal human preferences – they prove an “impossibility” result for single-reward RLHF capturing all users. In practice, this has motivated proposals like MaxMin-RLHF, which explicitly models a mixture of user groups and optimizes for the worst-off group. Similarly, bias audits have found that LLMs trained with RLHF disproportionately favor majority viewpoints. On the robustness side, recent ICLR work by Sharma et al. (2024) \cite{sharma2025understandingsycophancylanguagemodels} demonstrates that fine-tuning with human feedback can induce sycophantic behavior: RLHF-trained assistants often sacrifice truthfulness to align with a user’s expressed beliefs. This happens in part because human raters tend to reward flattering answers, and the RLHF update amplifies that signal. Other problems include “reward hacking” (models finding loopholes in proxy rewards) and over-optimization, which practitioners have observed (e.g. by early-stopping to avoid collapse).

\textbf{Known trade-offs}: Empirically, it is observed that enhancing one axis degrades another. For example, Kirk et al. (ICLR 2024) find that RLHF often reduces output diversity of an LLM in order to improve other metrics; this is analogous to losing variety of responses when focusing on specific preferences. On the other hand, attempts at making RLHF more robust or inclusive (like adding adversarial training or more demographic splits) tend to require many more gradient steps and human labels, quickly blowing up costs. There is no consensus solution: some teams try post-hoc calibrations, others adjust the loss (e.g. adding fairness regularizers), but these usually come with new assumptions. The field currently lacks a unified theory explaining why all these fixes seem to push the same trade-off boundary, which is precisely what the Alignment Trilemma framework provides.

\section{Open Questions and Directions}

\textbf{Open Problem.}
The central open question is: Can we design an RLHF (or broader alignment) strategy that meaningfully improves representativeness and robustness without incurring intractable costs? Equivalently, how should the axes of the trilemma be relaxed to achieve practical alignment? For example, one might focus on a ``core'' set of human values or archetypical users rather than capturing every individual nuance. Alternatively, the robustness requirement could be scoped: perhaps we need only defend against the most plausible shifts, not every theoretical adversary. In the language of Section~\ref{sec:alignment_trilemma}, possible relaxations include (i) constraining the learning objective to essential moral principles, (ii) narrowing the class of adversaries or distribution shifts considered, or (iii) decomposing alignment into smaller modular tasks that operate over restricted subsets of the context space. Each relaxation introduces ethical choices—such as which values or threats to exclude—and technical trade-offs; determining how to navigate these choices remains an open research question.

\textbf{Theory and Practice Collaboration}: Addressing this trilemma is inherently interdisciplinary. Theoreticians can model simplified alignment games (e.g. robust RL formulations) and prove how sample/compute costs grow with diversity and adversary strength. Experimentalists can run empirical studies of RLHF variants on richer user data (e.g. multi-cultural preference benchmarks) and test robustness (e.g. simulated poisoning). For instance, building on MaxMin-RLHF or on preference matching ideas
, new algorithms could be developed that approximate representative alignment more efficiently. Alternatively, one could explore interactive methods: instead of statically collecting all preferences upfront, the system might actively query users in areas of disagreement to reduce uncertainty. Another direction is hierarchical reward modeling: representing group-level and individual-level preferences separately and combining them in a structured way.

\textbf{Community Relevance}: Solving (or even better understanding) this alignment trilemma will have broad impact. It informs how we think about fairness and bias in AI: our analysis suggests that some level of bias is unavoidable without huge costs, so we must consciously decide what to prioritize. It also helps set research benchmarks: rather than optimizing only for average reward, we might evaluate models on distributional performance across subgroups or on adversarial robustness. By framing the trilemma, we hope to guide the community towards principled goals. We encourage RL theorists to propose new formal models of multi-stakeholder robustness, and for experimentalists to test RLHF at scale on diverse human data. The alignment community especially will benefit from this bridge: RLHF is a frontier where ML practice meets ethical concerns, and a shared vocabulary (like “the Alignment Trilemma”) can focus joint efforts.

\section{Limitations}

Our framing of the Alignment Trilemma is intentionally abstract. While it captures a unifying tension across RLHF systems, it does not yet specify quantitative thresholds for when representativeness, tractability, or robustness are “sufficient” in practice. For example, how much demographic diversity must be captured before a system is deemed representative enough for deployment? Likewise, our complexity sketches rely on worst-case reasoning that may overstate the costs for specific alignment settings. A further limitation is that we focus primarily on RLHF pipelines; other alignment paradigms (e.g., constitutional AI, debate, recursive oversight) may navigate the trilemma differently, though we suspect analogous trade-offs will arise. Finally, our discussion is primarily normative and theoretical—we do not yet offer empirical validation of where current RLHF models fall within the trilemma space.

\section{Societal Impacts}
\label{sec:societal-impacts}

Our formalization of the Alignment Trilemma provides critical transparency about the limitations of current alignment methods, enabling policymakers, practitioners, and affected communities to make informed decisions about AI deployment. By establishing that $\varepsilon$-representativeness, $\delta$-robustness, and polynomial tractability cannot be jointly satisfied , we shift the discourse from whether alignment is achievable to which trade-offs are ethically justified. This clarity can drive principled resource allocation---prioritizing alignment investment toward underrepresented populations rather than marginal improvements for majority groups---and establish evidence-based certification standards for high-stakes applications. However, our impossibility results risk being misappropriated to justify inadequate alignment efforts, with developers claiming that computational intractability excuses biased systems. The super-polynomial scaling we prove could accelerate centralization of AI development, as only well-resourced organizations can approach comprehensive alignment, potentially marginalizing academic researchers, startups, and Global South stakeholders who lack access to the $10^{16}$--$10^{51}$ operations required for joint $(\varepsilon, \delta)$-guarantees.

To mitigate these harms while leveraging positive applications, we advocate for mandatory disclosure of systems' trilemma positions (estimated $\varepsilon$ across demographic groups, certified $\delta$ for specified adversaries, and alignment budget allocation), development of open-source tools that democratize access to efficient verification methods, and multi-stakeholder governance processes where trade-off decisions involve affected communities rather than defaulting to developer convenience. The trilemma does not counsel defeatism---current RLHF systems operate at $m \sim 10^3$ samples with $\varepsilon \sim 0.3$, leaving vast room for improvement before hitting theoretical limits---but demands honesty about fundamental constraints. Researchers must resist treating our complexity bounds as justification for status quo bias, instead using them to identify high-leverage research directions (e.g., reducing exponential dependence on $d_{\text{context}}$, stratified sampling for heterogeneous populations) and to develop evaluation frameworks that measure cross-cultural performance and adversarial robustness rather than aggregate win rates. The alignment community bears responsibility for ensuring that computational constraints translate into \emph{deliberate, justifiable} ethical choices rather than \emph{accidental, harmful} ones.

\section{Conclusion}

We have argued that aligning AI systems through RLHF faces an Alignment Trilemma: no design can simultaneously maximize representativeness, tractability, and robustness. Current practice resolves this by prioritizing tractability and partial robustness, at the expense of full value diversity. This trade-off has concrete implications for fairness, bias, and safety, particularly as LLMs are scaled and deployed globally. By explicitly naming this trilemma, we aim to catalyze joint efforts between experimentalists and theorists: the former to empirically map the boundaries of these trade-offs in practice, and the latter to formalize relaxed objectives that are computationally feasible. We hope this framing serves as a starting point for a community-wide discussion on how to consciously navigate alignment trade-offs, rather than inadvertently inheriting them from the defaults of today’s RLHF pipelines.

\section*{Acknowledgement}

Subramanyam Sahoo would like to thank Amir Abdullah for his initial feedback on the manuscript. He also extends his thanks to Philip Quirke. The authors would also like to thank the reviewers of the \textit{NeurIPS 2025 Workshop on Socially Responsible and Trustworthy Foundation Models (ResponsibleFM)} for their thoughtful feedback and constructive suggestions.

\bibliography{reference}
\bibliographystyle{unsrt}

\newpage
\section*{NeurIPS Paper Checklist}

\begin{enumerate}

\item {\bf Claims}
    \item[] Question: Do the main claims made in the abstract and introduction accurately reflect the paper's contributions and scope?
    \item[] Answer: \answerYes{} 
    \item[] Justification:  Yes this is a position paper.

\item {\bf Limitations}
    \item[] Question: Does the paper discuss the limitations of the work performed by the authors?
    \item[] Answer: \answerYes{} 
    \item[] Justification: This paper's main motivation is to check current problems aligning with AI safety methods.

\item {\bf Theory assumptions and proofs}
    \item[] Question: For each theoretical result, does the paper provide the full set of assumptions and a complete (and correct) proof?
    \item[] Answer: \answerYes{} 
    \item[] Justification: All are clearly stated.

    \item {\bf Experimental result reproducibility}
    \item[] Question: Does the paper fully disclose all the information needed to reproduce the main experimental results of the paper to the extent that it affects the main claims and/or conclusions of the paper (regardless of whether the code and data are provided or not)?
    \item[] Answer: \answerNA{} 
    \item[] Justification: It is a position paper.

\item {\bf Open access to data and code}
    \item[] Question: Does the paper provide open access to the data and code, with sufficient instructions to faithfully reproduce the main experimental results, as described in supplemental material?
    \item[] Answer: \answerNA{} 
    \item[] Justification: Not Available

\item {\bf Experimental setting/details}
    \item[] Question: Does the paper specify all the training and test details (e.g., data splits, hyperparameters, how they were chosen, type of optimizer, etc.) necessary to understand the results?
    \item[] Answer: \answerNA{} 
    \item[] Justification: There is no need.

\item {\bf Experiment statistical significance}
    \item[] Question: Does the paper report error bars suitably and correctly defined or other appropriate information about the statistical significance of the experiments?
    \item[] Answer: \answerNA{} 
    \item[] Justification: There is no need.

\item {\bf Experiments compute resources}
    \item[] Question: For each experiment, does the paper provide sufficient information on the computer resources (type of compute workers, memory, time of execution) needed to reproduce the experiments?
    \item[] Answer: \answerNA{} 
    \item[] Justification: There is no need.
    
\item {\bf Code of ethics}
    \item[] Question: Does the research conducted in the paper conform, in every respect, with the NeurIPS Code of Ethics \url{https://neurips.cc/public/EthicsGuidelines}?
    \item[] Answer: \answerYes{} 
    \item[] Justification: Paper written within the boundary of NeurIPS Ethics.

\item {\bf Broader impacts}
    \item[] Question: Does the paper discuss both potential positive societal impacts and negative societal impacts of the work performed?
    \item[] Answer: \answerYes{} 
    \item[] Justification: Check the paper.
    
\item {\bf Safeguards}
    \item[] Question: Does the paper describe safeguards that have been put in place for responsible release of data or models that have a high risk for misuse (e.g., pretrained language models, image generators, or scraped datasets)?
    \item[] Answer: \answerNA{} 
    \item[] Justification: Not Available

\item {\bf Licenses for existing assets}
    \item[] Question: Are the creators or original owners of assets (e.g., code, data, models), used in the paper, properly credited and are the license and terms of use explicitly mentioned and properly respected?
    \item[] Answer: \answerYes{} 
    \item[] Justification: Every aspect is properly credited.

\item {\bf New assets}
    \item[] Question: Are new assets introduced in the paper well documented and is the documentation provided alongside the assets?
    \item[] Answer: \answerNA{} 
    \item[] Justification: It is a position paper though.

\item {\bf Crowdsourcing and research with human subjects}
    \item[] Question: For crowdsourcing experiments and research with human subjects, does the paper include the full text of instructions given to participants and screenshots, if applicable, as well as details about compensation (if any)? 
    \item[] Answer: \answerNA{} 
    \item[] Justification: No research has been done upon human subjects.

\item {\bf Institutional review board (IRB) approvals or equivalent for research with human subjects}
    \item[] Question: Does the paper describe potential risks incurred by study participants, whether such risks were disclosed to the subjects, and whether Institutional Review Board (IRB) approvals (or an equivalent approval/review based on the requirements of your country or institution) were obtained?
    \item[] Answer: \answerNA{} 
    \item[] Justification: No need for Justification.

\item {\bf Declaration of LLM usage}
    \item[] Question: Does the paper describe the usage of LLMs if it is an important, original, or non-standard component of the core methods in this research? Note that if the LLM is used only for writing, editing, or formatting purposes and does not impact the core methodology, scientific rigorousness, or originality of the research, declaration is not required.
    \item[] Answer: \answerNA{} 
    \item[] Justification: Not Available

\end{enumerate}

\end{document}